\documentclass[11pt]{article}\usepackage[]{graphicx}\usepackage[]{color}
\makeatletter
\def\maxwidth{ %
  \ifdim\Gin@nat@width>\linewidth
    \linewidth
  \else
    \Gin@nat@width
  \fi
}
\makeatother

\definecolor{fgcolor}{rgb}{0.345, 0.345, 0.345}

\usepackage{framed}
\makeatletter
 {\par\unskip\endMakeFramed%
 \at@end@of@kframe}
\makeatother

\definecolor{shadecolor}{rgb}{.97, .97, .97}
\definecolor{messagecolor}{rgb}{0, 0, 0}
\definecolor{warningcolor}{rgb}{1, 0, 1}
\definecolor{errorcolor}{rgb}{1, 0, 0}
\newenvironment{knitrout}{}{} % an empty environment to be redefined in TeX

\usepackage{alltt}

\usepackage{amsmath}
\usepackage{amssymb}
\usepackage{natbib}
\usepackage{graphicx}
\usepackage{algorithm2e}
\usepackage{xcolor}

\usepackage[T1]{fontenc}
\usepackage{lmodern}
\usepackage{bm}
\usepackage{acronym}
\usepackage[utf8]{inputenc}
\usepackage[english]{babel}
\usepackage{textcomp}

\usepackage{etoolbox}
\usepackage{siunitx}
\usepackage{microtype}
\usepackage{hyperref}
\usepackage{booktabs}

\usepackage[margin=1in]{geometry}

\let\oldFootnote\footnote
\newcommand\nextToken\relax

\renewcommand\footnote[1]{%
    \oldFootnote{#1}\futurelet\nextToken\isFootnote}

\newcommand\isFootnote{%
    \ifx\footnote\nextToken\textsuperscript{,}\fi}

\newif\ifisblinded
\isblindedtrue
\newcommand\doblind[2]{%
\ifisblinded %
#1 %
\else %
#2 %
\fi}

\SetKwInput{KwInput}{Input}
\SetKwInput{KwOutput}{Output}

\newcommand\bftab{\fontseries{b}\selectfont}
\robustify\bftab
\def\Cpp{{C\nolinebreak[4]\hspace{-.05em}\raisebox{.4ex}{\tiny\bf ++}}}

\newcommand\mat[1]{\mathbf{#1}}
\renewcommand\vec{\bm}
\newcommand\der{\operatorname{d\!}{}}
\newcommand\bigO[1]{\mathcal{O}\left(#1\right)}

\DeclareMathOperator\Prob{P}

\acrodef      {VA}{variational approximation}
\acrodefplural{VA}{variational approximations}

\acrodef      {GLMM}{generalized linear mixed model}
\acrodefplural{GLMM}{generalized linear mixed models}

\acrodef      {CDF}{cumulative density function}
\acrodefplural{CDF}{cumulative density functions}

\acrodef      {GSM}{generalized survival model}
\acrodefplural{GSM}{generalized survival models}

\acrodef      {PDF}{probability density function}

\acrodef      {MC}{Monte Carlo}

\acrodef      {KL}{Kullback–Leibler}

\acrodef      {GHQ}{Gauss–Hermite quadrature}

\acrodef      {AGHQ}{adaptive Gauss–Hermite quadrature}

\acrodef      {GWI}{Gaussian weighted integral}
\acrodefplural{GWI}{Gaussian weighted integrals}

\acrodef      {EM}{expectation maximization}

\acrodef      {AEM}{approximate expectation maximization}

\acrodef      {ECM}{expectation conditional maximization}

\acrodef      {GHK}{Geweke-Hajivassiliou-Keane simulator}

\acrodef      {MC}{Monte Carlo}

\acrodef      {QMC}{quasi-Monte Carlo}

\acrodef      {RQMC}{randomized quasi-Monte Carlo}

\acrodef      {IFM}{inference functions for margins}

\acrodef      {RMSE}{root mean square error}
\acrodefplural{RMSE}{root mean square errors}

\acrodef      {SVRG}{stochastic variance reduced gradient}

\acrodef      {GHK}{Geweke, Hajivassiliou, and Keane}

\acrodef      {MCMC}{Markov chain Monte Carlo}

\acrodef      {SMAE}{scaled mean absolute error}

\acrodef      {LEV}{lecturers evaluation}

\acrodef      {ESL}{employee selection}

\acrodef      {GBSG}{German breast cancer study group}

\acrodef      {TIPS}{restaurant tips}

\acrodef      {MLE}{maximum likelihood estimate}
\acrodefplural{MLE}{maximum likelihood estimates}

\acrodef      {SIMD}{single instruction, multiple data}

\acrodef      {TMVN}{truncated multivariate normal distribution}

\title{Asymptotically Exact and Fast Gaussian Copula Models for Imputation %
       of Mixed Data Types}

\author{Benjamin Christoffersen\thanks{%
   Department of Medical Epidemiology and Biostatistics, Karolinska Institutet, Sweden.}~\thanks{%
  Division of Robotics, Perception and Learning, KTH Royal Institute of Technology, Sweden.}~\thanks{%
  Swedish e-Science Research Center, Sweden.} \and
  Mark Clements\footnotemark[1]~\footnotemark[3] \and
  Keith Humphreys\footnotemark[1]~\footnotemark[3] \and
  Hedvig Kjellström\footnotemark[2]~\footnotemark[3]}
\IfFileExists{upquote.sty}{\usepackage{upquote}}{}
\begin{document}

\maketitle

\begin{abstract}
Missing values with mixed data types is a common problem in
a large number of machine learning applications
such as processing of surveys
and in different medical applications.
Recently, Gaussian copula models have been
suggested as a means of performing imputation of missing values
using a probabilistic
framework. While the present Gaussian copula models have shown to yield
state of the art performance, they have two limitations:
they are based on an approximation
that is fast but may be imprecise and they do not support unordered
multinomial variables.
We address the first limitation using direct and arbitrarily precise
approximations both for model estimation and imputation by using
randomized quasi-Monte Carlo procedures.
The method we provide has lower errors for the estimated model parameters
and the imputed values, compared to previously proposed methods.
We also extend the
previous Gaussian copula models to include unordered multinomial variables
in addition to the present support of ordinal, binary, and continuous variables.

~\\[-.8em]

\end{abstract}

\section{Introduction}

Data sets in medical applications, surveys, user ratings, etc.\
are becoming larger which brings possibilities for machine
learning applications. However, these larger data sets often have missing
values. Therefore, imputation of missing values becomes increasingly important
as a preprocessing step to many applications.
\cite{zhao20,zhao20Mat,landgrebe20} describe a method for imputation for
continuous, binary,
and ordinal variables using Gaussian copulas which yields state of the art
performance. They address the important issue
of performing imputation for data sets containing variables such as
ratings in reviews, integer scales of how severe the spread of a tumor is
(medical data), and rank variables in surveys, which need to
be analyzed in combination with continuous
variables such as age, income, etc.
using a Gaussian copula.
This is straightforward since it is often easy
to describe the marginal distribution of each variable,
few assumptions are made about the
marginals, and use of such a probabilistic framework allows for
construction of confidence intervals for the imputed values
similar to the measure developed
by \cite{zhao20Mat}. Lastly,
the Gaussian copulas have computational advantages.

However, there are three open issues. Firstly, \cite{zhao20} use an
\ac{AEM} algorithm in a way that is
related to the work by \cite{Guo15}.
While the approximation is fast, the approximation may yield inefficient
and possibly even biased results. Secondly,
like \ac{IFM} for fully parametric models,
they use a two-stage estimation method
that may be inefficient in some cases. Finally, their model does not
support multinomial variables.\footnote{%
We will write multinomial when we refer to unordered multinomial and ordinal
when we refer to ordered multinomial.}
Thus, for many data sets, their method cannot be used.

This paper makes two main contributions:

\begin{enumerate}
  \item \label{enum:newEst}
   The method we provide gives an asymptotically exact and fast approximation of
     the of the log marginal likelihood, the derivatives, and the quantities
     needed to perform imputation.
   Moreover, we estimate some of the parameters in the marginal distributions
     jointly with the copula parameters, instead
     of using an \ac{IFM} like method, which increases the efficiency.

  Our method provides
     conditional probabilities for missing binary, ordinal, and
     multinomial variables with arbitrary precision. The
     \ac{AEM} method cannot provide such approximations
     which are key for multiple imputation. This contribution is mainly
     described in Section \ref{sec:newMethod} and partly in Section
     \ref{sec:addMult}.
  \item \label{enum:AddMult}
   Our Gaussian copula model
   supports multinomial variables in addition to binary, ordinal, and continuous
     variables. Thus, our method is applicable to a large number of data sets
     with mixed data types. This is further described in Section
     \ref{sec:addMult}.
\end{enumerate}

We will cover the Gaussian copula model
in Section \ref{sec:GaussianCopula} but start with a review of related work in Section \ref{sec:related}.
\doblind{%
Our software will be made publicly available and is provided along with
the script to produce the results in the supplementary material.%
}{%
All our software is freely available in an open source R package named
mdgc which can be installed from
\url{https://github.com/boennecd/mdgc}. The implementation is mainly written
in \Cpp ~and supports computation in parallel.
}

\section{Related Work}\label{sec:related}
\cite{zhao20} method was based on earlier work on Gaussian copula models
\citep{Hoff07,Liu09,Murray13,Fan17,Feng19,Cui19}. These models can be
seen as a generalization of the linear mixed model in the sense that
we no longer assume that the marginals for the continuous variables are normally
distributed. Instead we assume that they have been transformed with
bijective transformations and, thus,
allow for greater flexibility for the marginal distributions.
Secondly, there are binary and ordinal variables that are
created by cutting the latent variables into bins.
The resulting model can be estimated with
semiparametric methods. However, unlike in the linear mixed model,
there is no closed form solution for the likelihood.

Previous work on related models has
focused on model estimation rather than imputation.
Moreover, some methods are only
able to estimate the model with complete data. The \ac{MCMC} method suggested
by \cite{Hoff07} is an exception but it can be very slow. Thus, \cite{zhao20}
propose to estimate the model parameters with a frequentist approach.
Their estimation method and imputation method is based on an \ac{EM} algorithm
which requires evaluation of moments of the \ac{TMVN}.
They approximate the moments using an approximation similar to the one
suggested by \cite{Guo15}.
This approximation is fast and, as \cite{zhao20} show, it yields
superior single imputation performance, in their examples, compared with
state of the art non-parametric methods such as the random forest based
missForest \citep{Stekhoven12} and the principal components based
imputeFAMD \citep{Audigier14,Josse16}.

Although \cite{zhao20Mat} state that direct maximum likelihood is hard to
optimize because the likelihood involves Gaussian integrals,
moderately precise and, importantly, fast \ac{RQMC} procedures have been
developed by \cite{Genz92,Hajivassiliou96,Genz02}. These methods are also easy
to generalize to related quantities like derivatives and conditional means
or probabilities for missing values. Having arbitrarily precise procedures are
important as they can yield more efficient estimators of the model parameters,
which the researcher may be interested in per se, and can potentially improve
the imputation.

\section{Gaussian Copula Models}\label{sec:GaussianCopula}

We will cover the model that \cite{zhao20} use, and thereafter
provide our extension to include
multinomial variables and alternative methods for estimation and imputation.
$\vec X_i$ denotes the vector with $K$
variables for observation
$i=1,\dots,n$ where some entries
may be missing. The $X_{ij}$'s for
$j\in \mathcal C \subseteq \{1,\dots K\}$ are continuous,
the entries with $j\in \mathcal O \subseteq \{1,\dots K\}$, with
$\mathcal O \cap \mathcal C = \emptyset$, are ordinal,
and the entries with
$j\in \mathcal B =  \{1,\dots K\} \setminus \mathcal O \cup \mathcal C$
are binary.
For now, we assume that all entries are observed
with value $\vec x_1,\dots, \vec x_n$.

Let $\Phi^{(k)}(\vec a, \vec b;\vec\mu, \mat\Sigma)$ be the
$k$-dimensional multivariate normal distribution \ac{CDF} with mean
$\vec\mu$ and covariance matrix $\mat\Sigma$ over the
box with limits at $\vec a$ and $\vec b$ given by%
$$
\Phi^{(k)}(\vec a, \vec b;\vec\mu, \mat\Sigma) =
 \int_{a_1}^{b_1}\cdots\int_{a_k}^{b_k}
 \phi^{(k)}(\vec u;\vec\mu,\mat\Sigma)\der \vec u
$$%
where $\phi^{(k)}$ is the corresponding density function.
We omit the mean vector and covariance matrix in the
standard case, omit the
superscript in the univariate case, and write
$\Phi^{(k)}(\vec b;\vec\mu, \mat\Sigma)$ when
$\vec a = (-\infty,\dots,-\infty)^\top$.
\cite{zhao20,zhao20Mat} use a Gaussian copula model where
it is assumed that there
is a $K$-dimensional latent variable $\vec Z_i \in\mathbb R^K$ such that
$\vec Z_i \sim N^{(K)}\left(\vec 0_K,\mat\Sigma\right)$
where $\vec 0_k$ is a vector with $k$ zeros and %
$$
\mat\Sigma = \begin{pmatrix}
    1 & \sigma_{12} & \cdots & \sigma_{1K} \\
    \sigma_{21} & 1 & \ddots  & \vdots \\
    \vdots & \ddots & \ddots & \sigma_{K - 1, K} \\
    \sigma_{K2} & \dots & \sigma_{K, K - 1} & 1
  \end{pmatrix}.
$$%
The relation to the observed outcomes is%
\begin{align}
X_{ij} &= f_j(Z_{ij}),\quad j \in \mathcal C \nonumber\\
X_{ij} &= \begin{cases}
  1 & Z_{ij} > -\Phi^{-1}(p_{j}) \\
  0 & \text{otherwise}  \label{eqn:BinarySpec}\\
\end{cases}, \quad j \in \mathcal B \\
X_{ij} &= k\Leftrightarrow \alpha_{jk} < Z_{ij} \leq \alpha_{j,k + 1},
  \quad
   j \in \mathcal O\ \label{eqn:OrdinalSpec}
\end{align}%
where $k = 0,\dots m_j -1$ in Equation \eqref{eqn:OrdinalSpec},
$f_j$ is a given unknown bijective function if the $j^{\text{th}}$ variable is continuous,
$p_j$ is the unknown marginal
probability of $X_{ij} = 1$ if the $j^{\text{th}}$ variable is binary,
$m_j$ is the number of
categories of the $j^{\text{th}}$ variable if it is ordinal, and
$\alpha_{j0}, \dots,  \alpha_{jm_j}$ are bounds for the
$j^{\text{th}}$ variable if it is ordinal with $\alpha_{j0} = -\infty$
and $\alpha_{jm_j} = \infty$.

The interpretation of the models is that
the continuous variables are transformed normally
distributed variables, the binary variables are thresholded
normally distributed variables, and the ordinal variables are
normally distributed variables which are cut into bins.
The flexibility of the Gaussian copula models is due to the few
assumptions about the $f_j$'s for the continuous variables. Thus, the marginal
distribution for each variable can be very complex. The parametric assumption
is the particular copula we use. Other copulas can be used
but the Gaussian copula has computational advantages which we extensively
use in Sections \ref{sec:newMethod} and \ref{sec:addMult}.

\cite{zhao20,zhao20Mat}
fit the model by first estimating the marginals,
i.e.\ estimating the $f_j$'s for the continuous variables
using rescaled empirical \acp{CDF}, estimating the $p_j$'s for the binary
variables, and estimating the borders
$\alpha_{j1},\dots,\alpha_{j,m_j - 1}$ for the ordinal variables.
They then
estimate $\mat\Sigma$ conditional on the parameters for the
marginal distributions. This two-stage estimation method is often
referred to as \ac{IFM} in the fully parametric case where there are no
continuous variables.

\cite{Joe05} shows that \ac{IFM} has a high
relative efficiency compared with the \ac{MLE}
of all parameters
(that is, joint estimation of the marginals and $\mat\Sigma$)
in related models. However, the efficiency tends to decrease as
the number of variables increase or when the dependence is high.
Therefore, we perform joint estimation of some of the marginal
distribution parameters which \cite{zhao20,zhao20Mat} fix in
the second step of \ac{IFM}. In particular, we can let
$Z_{ij}$ for a binary variable have a non-zero mean given by
$\mu_j = \Phi^{-1}(p_{j})$ and assume that $X_{ij} = 1$ if
$Z_{ij} > 0$. We then jointly estimate $\mu_j$ and $\mat\Sigma$ for
$j\in\mathcal B$. We still use a two-step procedure where we fix the borders
for the ordinal variables (the $\alpha$s) and estimate the $f_j$'s
non-parametrically.\footnote{
It is also possible to jointly estimate the borders for the ordinal
variables and the $f_j$'s if we parameterize them. We discuss this further in
the discussion in Section \ref{sec:conclusion}.}

Let
$\vec X_{i\mathcal I} = (x_{l_1},\dots, x_{l_k})^\top$
where $\mathcal I = \{l_1,\dots, l_k\}$
and let $\hat f_j$ be the estimate of $f_j$. Let
$\widehat{\vec z}_{i\mathcal C}$ be a vector with
$\widehat{\vec z}_{ij} = \hat f_j^{-1}(x_{ij})$. Then
the log marginal likelihood conditional on the estimated
marginals is
$l(\mat\Sigma,\vec\mu) = \sum_{i = 1}^n l_i(\mat\Sigma,\vec\mu)$ where%
\begin{align}
l_i(\mat\Sigma,\vec\mu) =
  \log\phi^{(\lvert\mathcal C\rvert)}(
  \widehat{\vec z}_{i\mathcal C}; \vec 0,
  \mat\Sigma_{\mathcal C\mathcal C}) +
  \log \Phi^{(\lvert\mathcal B\cup\mathcal O\rvert)}
  \bigg(&\vec a_i,\vec b_i;
  \vec\mu_{\mathcal B\cup\mathcal O,\mathcal C} +
  \mat\Sigma_{\mathcal B\cup\mathcal O,\mathcal C}
  \mat\Sigma_{\mathcal C\mathcal C}^{-1}
  \widehat{\vec z}_{i\mathcal C},
  \label{eqn:orgLogML}\\
& \mat\Sigma_{\mathcal B\cup\mathcal O,\mathcal B\cup\mathcal O} -
  \mat\Sigma_{\mathcal B\cup\mathcal O,\mathcal C}
  \mat\Sigma_{\mathcal C, C}^{-1}
  \mat\Sigma_{\mathcal C, \mathcal B\cup\mathcal O}
  \bigg).
  \nonumber
\end{align}%
$\vec a_i$ and $\vec b_i$ depend on the estimated
borders for the ordinal variables along with the
observed variables, $\vec x_{\mathcal B\cup\mathcal O}$,
as explained by \cite{zhao20}, and
$\vec\mu_{\mathcal B\cup\mathcal O,\mathcal C}$ contains the
possibly non-zero means for the binary variables. The $\vec a_i$ and $\vec b_i$
entries for the binary (and later multinomial variables) are $-\infty$ and 0
or 0 and $\infty$, respectively.

\section{New Estimation and Imputation Method}\label{sec:newMethod}
The main computational burden in evaluating the log marginal likelihood
in Equation \eqref{eqn:orgLogML}
is to approximate the $\lvert\mathcal B\cup\mathcal O\rvert$ dimensional
\ac{CDF}. \cite{zhao20Mat} state that direct optimization is hard because of
the \ac{CDF} and, therefore, use an approximation of the type suggested by
\cite{Guo15} in an \ac{AEM} algorithm.
However,
\cite{Genz92,Genz02,Genz09} show
that the \ac{CDF} $\Phi^{(k)}(\vec a, \vec b;\vec\omega, \mat\Omega)$
can be approximated quickly using importance sampling using
the importance distribution with density $h$ given by %
\begin{align}
h(\vec u) &= \prod_{j = 1}^k \begin{cases}
   \phi(u_j) / w_j(\vec u) &
     \hat a_j(\vec u_{1:{(j - 1)}}) < u_j < \hat b_j(\vec u_{1:{(j - 1)}}) \\
   0 & \text{otherwise}
   \end{cases}\label{eqn:importDist}\\
w_j(\vec u) &= \Phi(
    \hat b_j(\vec u_{1:{(j - 1)}})) -
    \Phi(\hat a_j(\vec u_{1:{(j - 1)}}))
    \nonumber\\
\hat b_j(\vec x) &= \begin{cases}
   o_{11}^{-1}(b_1 - \omega_1) & j = 1 \\
   o_{jj}^{-1}(b_j - \omega_j - \vec o_{1:(j - 1),j}^\top\vec x) & j > 1
  \end{cases}
  \nonumber\\
\hat a_j(\vec x) &= \begin{cases}
   o_{11}^{-1}(a_1 - \omega_1) & j = 1 \\
   o_{jj}^{-1}(a_j - \omega_j - \vec o_{1:(j - 1),j}^\top\vec x) & j > 1
  \end{cases} \nonumber
\end{align}
where $\mat O$ is a Cholesky decomposition of $\mat\Omega$ such
that $\mat O^\top\mat O = \mat\Omega$. Thus, we use a method,
which is a \ac{MC} approximation of%
$$
\Phi^{(k)}(\vec a, \vec b;\vec\omega, \mat\Omega) =
  \int
  h(\vec u) \prod_{j=1}^kw_j(\vec u)\der \vec u.
$$%
\cite{Genz92,Genz02} use
a heuristic variable reordering to reduce the variance of the
estimator at a fixed number of samples and use \ac{RQMC} to get
a better bound on the error.
The advantage of using \ac{RQMC} is that the error is bounded by
$\bigO{s^{-1 + \epsilon}}$, or more precisely $\bigO{s^{-1}(\log s)^l}$ for some
$l\leq k$, where $s$ is the number samples \citep{caflisch98}.
This is in contrast to
the $\bigO{s^{-1/2}}$ bound of \ac{MC} methods.

Gradient approximations with respect to the mean and covariance
matrix of the log of the CDF can be written as
\begin{equation}\label{eqn:grad}
\frac{\int
  \vec g(\mat O^\top\vec u + \vec\omega;\vec\omega,\mat\Omega)
  h(\vec u) \prod_{j = 1}^k w_j(\vec u)\der \vec u}
{\Phi^{(k)}(\vec a, \vec b;\vec\omega, \mat\Omega)}
\end{equation}%
for a given function $\vec g$
as described by \cite{Hajivassiliou96} for the \ac{GHK} simulator they use.
This is the expectation of $\vec g(\vec X)$ where $\vec X$ follows a
\ac{TMVN} with location parameter $\vec\omega$,
scale parameter $\mat\Omega$, and truncated such that
$a_j < X_j < b_j$ for $j = 1,\dots,k$.
We have rewritten the
Fortran
code by \cite{Genz02,mvtnorm} in \Cpp~to also provide an approximation
to the numerator in Equation \eqref{eqn:grad}. Details of the method are
provided in supplementary material  \ref{sec:QMCMethod}.
Standard applications of
the multivariate version of the chain rule can then be used to get an
approximation of the gradient of the log marginal likelihood in
Equation \eqref{eqn:orgLogML}, once a gradient approximation
of the \ac{CDF} is implemented. Details are provided in supplementary material
\ref{sec:Derivs}.
The computational complexity of all our approximations are
$\bigO{nK^3}$ at a fixed number of \ac{RQMC} samples like the \ac{AEM} method.

The model can be estimated by using a log Cholesky
decomposition \citep{Pinheiro96} of $\mat\Sigma$.
Stochastic gradient descent methods are easy to apply,
because of the $n$ independent log marginal likelihood
terms, if one re-scales $\mat\Sigma$ to be a correlation matrix between
each iteration, as \cite{zhao20} do.
We have tried ADAM \citep{Kingma15} and \ac{SVRG} \citep{Johnson09}.
The latter seems to work well with an
appropriate learning rate. We have also implemented an augmented
Lagrangian method to avoid the ad hoc re-scaling.
On average, the augmented Lagrangian method
tends to provide slightly better estimates of $\mat\Sigma$
than \ac{SVRG}. We however omit this from our comparisons
because it is slower.

For the imputations, we use the conditional means for the
latent variables (the $Z_{ij}$'s) corresponding to
missing continuous variables (a missing variable with $j\in\mathcal C$)
and map back using $\hat f_j$'s.
This is similar to \cite{zhao20},
but our method can yield an arbitrarily precise
approximation of the conditional means on the latent
scale.
We use either conditional probabilities or medians
for each of the binary, ordinal,
and later multinomial variables.
These quantities are given by
a suitable choice of $\vec g$ in Equation \eqref{eqn:grad}.\footnote{%
The means can be computed by using the identity function
and the conditional probabilities can be computed with
a function which returns an one-hot vector which has a one in the category that
the sampled $\vec z_i$ implies. The conditional median for the
ordinal variables can be computed from the conditional probabilities.}

Thus, we are also able to approximate the quantities needed
for imputations with the new \Cpp
~code for \ac{CDF} approximation. \cite{zhao20} transform back
their approximate conditional means for the binary and ordinal variables.
Their method cannot
directly be used to provide conditional probabilities.
Finally, as \cite{zhao20,zhao20Mat} do, we assume that data is missing
completely at random
and leave handling of data which is missing at random for future work.

As Equation \eqref{eqn:grad} is the expectation of
$\vec g(\vec X)$ where $\vec X$ follows a \ac{TMVN}, one could directly
sample from the \ac{TMVN} to avoid the
separate computation of the denominator and numerator. However, sampling from
a \ac{TMVN} is hard. Interestingly, \cite{Botev17} has recently extended the
work by \cite{Genz92,Genz02} to sample from a \ac{TMVN} using an
accept-reject sampling schema based on a minimax tilting method with a tilted version
of Equation \eqref{eqn:importDist}. It will be interesting
to apply the method by \cite{Botev17} in our application but this is beyond the
scope of the present paper.

\subsection{Application of the New Method}\label{sec:compWOMult}

% latex table generated in R 4.1.0 by xtable 1.8-4 package
% Thu Jul  1 11:57:47 2021
\begin{table*}[t]
\centering
\caption{Mean classification error, \acs{RMSE}, and \acs{SMAE} each with plus or minus two standard errors for the first simulation study. The error rows are the mean classification errors for the ordinal variables and the mean \acsp{RMSE} of the continuous variables. The last two columns show the mean computation times in seconds and relative errors for the correlation matrix estimates. The best result in each comparison is in bold.} 
\label{tab:simZhaoError}
\begingroup\tiny
\begin{tabular}{llS[table-format = 1.4(4)]S[table-format = 1.3(3)]S[table-format = 1.3(3)]|S[table-format = 3(3)]S[table-format = 1.4(4)]}
  \toprule
Metric & Method & {Binary} & {Ordinal} & {Continuous} & {Time} & {Relative $\mat\Sigma$ error} \\ 
  \midrule
Error & RQMC (our) & \bftab 0.2429(9) & \bftab 0.582(1) & \bftab 0.741(2) & \bftab 184(4) & \bftab 0.0812(4) \\ 
   & Median (our) &  & 0.602(1) &  &  &  \\ 
   & AEM (ZU) & 0.2502(9) & 0.615(1) & 0.750(2) & 412(2) & 0.1172(6) \\ 
  SMAE & RQMC (our) & \bftab 0.4861(18) & 0.694(3) & \bftab 0.705(2) &  &  \\ 
   & Median (our) &  & \bftab 0.650(2) &  &  &  \\ 
   & AEM (ZU) & 0.5008(18) & 0.659(2) & 0.710(2) &  &  \\ 
   \bottomrule
\end{tabular}
\endgroup
\end{table*}

\begin{knitrout}
\definecolor{shadecolor}{rgb}{0.969, 0.969, 0.969}\color{fgcolor}\begin{figure}[t!]
\includegraphics[width=\maxwidth]{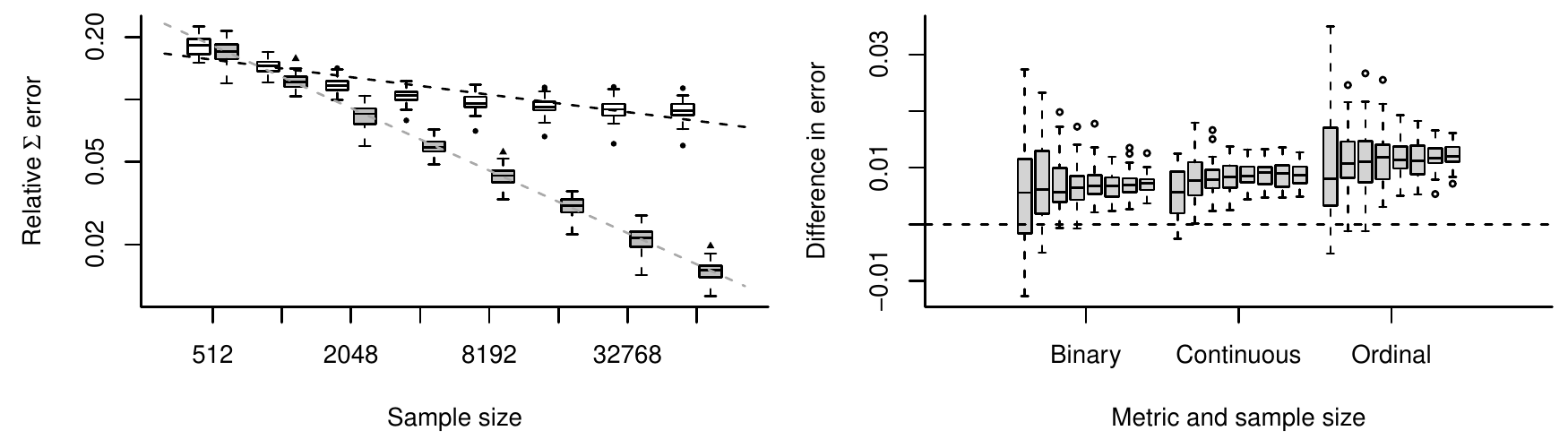} \caption{The left plot shows the relative error of the covariance matrix versus the sample size on the log-log scale. The gray boxes are the \acs{RQMC} method and the white boxes are the \acs{AEM} method. Regression lines are added on the log-log scale. The right plot shows the difference in mean RMSE and classification error for the three types of data for increasing sample sizes (increasing from left to right). We use the median method for the ordinal outcomes with \acs{RQMC} as this is the closest to the \acs{AEM} method.}\label{fig:simZhaoRngVcovVaryNObs}
\end{figure}

\end{knitrout}

We now address how our new estimation method and imputation method
compare with the \ac{AEM} based method of
\cite{zhao20}, when (i) the model is correctly specified, and (ii) when the
methods are applied to observational data.
We start with (i).
An argument for using the \ac{AEM} method is that it is fast even when there
are moderately many variables (moderately large $K$) and many observations
(large $n$). We draw $n = 10000$ observations with $K = 60$
variables of
which  $20$ each are continuous, binary, and ordinal. As in  \cite{zhao20},
we use five equally likely categories
for the ordinal variables and generate $100$ data sets
where we mask each variable independently at random with a 30 percent chance.
For each data set, we sample
$\mat\Sigma$ from a Wishart distribution with $K$ degrees of freedom with an
identity matrix as the
scale matrix and scale the sampled matrix such that it is
a correlation matrix. Thus, the methods are compared on different correlation
matrices.
As in \cite{zhao20}, we let the continuous variables have
standard exponential (marginal) distributions.

We use three different metrics to measure the performance of the methods.
The first two
are also used by \cite{zhao20}. The user
may be interested in $\mat\Sigma$ per se and, thus, we use the relative error
$\lVert\widehat{\mat\Sigma} - \mat\Sigma\rVert_F / \lVert\mat\Sigma\rVert_F$,
where $\widehat{\mat\Sigma}$ is an estimated correlation matrix and
$\lVert\cdot\rVert_F$ is the Frobenius norm.
We use the \ac{SMAE} given by
$\sum_{i\in \mathcal H_j}\lVert \widehat X_{ij} - X_{ij}\rVert_1 /
  \sum_{i\in \mathcal H_j}\lVert \tilde X_{j} - X_{ij}\rVert_1$ where
$\mathcal H_j$ is the set of indices with missing entries for variable $j$,
$\widehat X_{ij}$ is an imputed value, and
$\tilde X_{j}$ is the median estimate based on the observed values.
Finally, we use the classification error for the binary and ordinal variables
and the \ac{RMSE} for the continuous variables.

Table \ref{tab:simZhaoError} summaries the results of the simulation study.
The \ac{RQMC} rows are our method which performs imputation by assigning
the category with the highest conditional probability.
The ``Median'' rows also use our estimation method but
selects the conditional median category similar to the
\ac{AEM} algorithm which back transforms an approximate mean.
Our new method performs better on all metrics.
Selecting the category with the highest conditional probability leads to a lower
classification error but higher \ac{SMAE}.

The computation time of our method is competitive although we stress that
the \ac{AEM} implementation does not support computation in parallel.\footnote{%
Our simulations are run on
a Laptop with a Intel\textsuperscript{\tiny\textregistered}~Core\texttrademark
~i7-8750H CPU,
our code is compiled with GCC 10.1.0, and we use four threads.
We use between 500 and 10000 samples per log marginal likelihood term
evaluation and 2000 to 20000 samples for the imputations.
The \ac{RQMC} approximations are stopped early if the relative error is small.
The method by
\cite{zhao20} is installed from \url{https://github.com/udellgroup/mixedgcImp}.}
Importantly, the average relative error for the correlation matrix is much
lower.

% latex table generated in R 4.1.0 by xtable 1.8-4 package
% Thu Jul  1 11:57:52 2021
\begin{table*}[t]
\centering
\caption{Mean classification error, \acp{RMSE}, and \acp{SMAE} (see Table \ref{tab:simZhaoError}) with four different data sets used by \cite{zhao20} and the two largest data sets from the catdata package. The average computation times in seconds are shown in the last column.} 
\label{tab:ZhaoCompare}
\begingroup\tiny
\begin{tabular}{lllS[table-format = 1.4(4)]S[table-format = 1.4(4)]S[table-format = 1.3(3)]|S[table-format = 2.2(2)]}
  \toprule
Data set & Metric & Method & {Binary} & {Ordinal} & {Continuous} & {Time} \\ 
  \midrule
Rent & Error & RQMC (our) & \bftab 0.1212(8) & \bftab 0.3702(39) & 0.747(4) & \bftab 31.67(20) \\ 
   &  & Median (our) &  & 0.3710(41) &  &  \\ 
   &  & AEM (ZU) & 0.1237(8) & 0.3716(39) & \bftab 0.746(4) & 35.41(65) \\ 
   & SMAE & RQMC (our) & \bftab 0.9721(48) & 0.5102(53) & \bftab 0.701(3) &  \\ 
   &  & Median (our) &  & \bftab 0.5095(54) &  &  \\ 
   &  & AEM (ZU) & 0.9974(16) & 0.5107(52) & 0.706(2) &  \\ 
   [.35em] Medcare & Error & RQMC (our) & \bftab 0.3418(16) & \bftab 0.4398(10) & \bftab 0.984(6) & \bftab 6.74(27) \\ 
   &  & Median (our) &  & 0.4661(10) &  &  \\ 
   &  & AEM (ZU) & 0.3466(17) & 0.4647(10) & 0.987(6) & 22.63(11) \\ 
   & SMAE & RQMC (our) & \bftab 0.8020(46) & 0.9823(11) & \bftab 0.951(1) &  \\ 
   &  & Median (our) &  & \bftab 0.9639(13) &  &  \\ 
   &  & AEM (ZU) & 0.8138(38) & 0.9641(10) & 0.952(1) &  \\ 
   [.35em] ESL & Error & RQMC (our) &  & 0.4911(36) &  & \bftab 1.80(31) \\ 
   &  & Median (our) &  & \bftab 0.4908(36) &  &  \\ 
   &  & AEM (ZU) &  & 0.4919(35) &  & 2.05(4) \\ 
   & SMAE & RQMC (our) &  & 0.5309(53) &  &  \\ 
   &  & Median (our) &  & \bftab 0.5238(53) &  &  \\ 
   &  & AEM (ZU) &  & 0.5254(52) &  &  \\ 
   [.35em] LEV & Error & RQMC (our) &  & \bftab 0.6337(23) &  & \bftab 2.20(33) \\ 
   &  & Median (our) &  & 0.6476(21) &  &  \\ 
   &  & AEM (ZU) &  & 0.6484(21) &  & 5.40(7) \\ 
   & SMAE & RQMC (our) &  & 0.9637(67) &  &  \\ 
   &  & Median (our) &  & \bftab 0.8696(28) &  &  \\ 
   &  & AEM (ZU) &  & 0.8714(27) &  &  \\ 
   [.35em] GBSG & Error & RQMC (our) & 0.2895(31) & 0.3610(44) & \bftab 0.921(8) & \bftab 2.64(35) \\ 
   &  & Median (our) &  & 0.3604(44) &  &  \\ 
   &  & AEM (ZU) & \bftab 0.2889(31) & \bftab 0.3572(46) & 0.925(8) & 7.13(15) \\ 
   & SMAE & RQMC (our) & 0.7276(84) & 1.0207(60) & 0.879(3) &  \\ 
   &  & Median (our) &  & 1.0189(61) &  &  \\ 
   &  & AEM (ZU) & \bftab 0.7257(82) & \bftab 1.0085(41) & \bftab 0.878(2) &  \\ 
   [.35em] TIPS & Error & RQMC (our) & \bftab 0.2857(58) & \bftab 0.4133(72) & 0.786(16) & 1.67(17) \\ 
   &  & Median (our) &  & 0.4414(88) &  &  \\ 
   &  & AEM (ZU) & 0.2874(62) & 0.4489(91) & \bftab 0.784(16) & \bftab 1.56(4) \\ 
   & SMAE & RQMC (our) & \bftab 0.8241(142) & 0.8007(174) & 0.763(8) &  \\ 
   &  & Median (our) &  & \bftab 0.7786(154) &  &  \\ 
   &  & AEM (ZU) & 0.8284(139) & 0.7881(155) & \bftab 0.761(8) &  \\ 
   \bottomrule
\end{tabular}
\endgroup
\end{table*}

For our second evaluation,
we compare the methods on six data sets.
We use four of the data sets where \cite{zhao20} show that
the \ac{AEM} method yields superior single imputation performance
compared to state of the art non-parametric methods. These are
the \ac{LEV} data set, \ac{ESL} data set, \ac{GBSG} data set, and \ac{TIPS}.
We also add the two largest
data sets from the catdata package \citep{Fox20}. This is the medcare data set
containing the number of physician office visits and
the rent in Munich data set.\footnote{%
We remove the rent per square meter because of the deterministic relationship
with rent and the living space and we remove the municipality
code as it is a multinomial variable.}
Both data sets have variables of all types (binary, ordinal, and continuous).
We generate $100$ data sets for each of the six
where we independently at random
mask each variable with a 30 percent chance and standardize all continuous variables
such that the \ac{RMSE} is not dominated by a few variables.
The first four data sets have between
$244$ and
$1000$
observations while the latter
two have
$2053$ and
$4406$ observations. Further details
about all data sets in this paper can be found in supplementary material  \ref{sec:dataInfo}.

Table \ref{tab:ZhaoCompare} summarises the results. With respect to the standard
errors there are only noticeable differences in the average errors for the
two larger data sets where our method is preferred.
A reason for there seemingly being no differences for the smaller
data sets can be that the \ac{AEM} method tends to select
a correlation matrix close to a diagonal matrix. The average differences between
the Frobenius norm of the estimated correlation matrix of the two methods
ranged from
$0.0123$ to $0.691$
over the six data sets.
This might lead to bias-variance trade-off.

To explore this further, we have
conducted a new simulation study with $K = 15$ variables
and $5$ of each type of outcome to reduce
the computation time. We use
$n = 512, 1024, \dots, 65536$
observations with 50 data sets for each $n$.
The results are summarized in Figure \ref{fig:simZhaoRngVcovVaryNObs}
which shows that the \acs{AEM} method does not improve in terms of
the relative error of the estimated covariance matrix (or at best very slowly)
for large sample sizes.
This makes it hard to believe that the \acs{AEM} method will converge, which
implies that it must be biased for some sample
sizes. This is troubling if the researcher is interested in the
covariance matrix, as in Figure 6 in \cite{zhao20short}.
In contrast, our proposed method has a regression slope of
-0.5006
consistent with the typical $\sqrt n$ convergence rate of maximum likelihood
estimators. Figure \ref{fig:simZhaoRngVcovVaryNObs} also shows that the
gap in imputation error increases as a function of the sample size.

To summarize, we find large improvements when the model is correctly specified and
only improvements or identical results with observational data sets.
There is a big
difference in the average error for the estimated correlation matrix for the
correctly specified model which is very important for researchers who are interested in the
correlation matrix per se. Importantly, the differences in the error of the two
methods increases as the number of samples increase
which is consistent with the observational data. This suggests
that the \ac{RQMC} is preferable when there are many observations.

\section{Adding Multinomial Variables}\label{sec:addMult}
We extend the model in this section to also support
multinomial variables.
To build up the intuition, we first show the formulation with a
simple model with
one binary variable $X_1 \in \{0, 1\}$ and
one continuous variable $X_2$. Suppose that %
\begin{align}
\vec Z & \sim N^{(3)}\left(
  \begin{pmatrix} 0 \\ \mu_2 \\ 0 \end{pmatrix},
  \begin{pmatrix}
    \xi & 0 & 0 \\
    0 & 1 & \sigma_{23} \\
    0 & \sigma_{32} & 1
  \end{pmatrix}
  \right) \label{eqn:Z3DMod}\\
X_1 &= j - 1 \Leftrightarrow Z_j \geq
  \max (Z_1, Z_2) \qquad j \in \{1,2\}
  \label{eqn:Z3DModX1}\\
X_2 &= f(Z_3) \nonumber
\end{align}%
for a bijective function $f$ and
$\mu_2 = \Phi^{-1}(p_1)$. Using that
$\vec Z_{1:2} \mid Z_3 = z \sim N(\vec m, \mat S)$ with
$\vec m = (0, \mu_2  + \sigma_{23}z)^\top$ and%
$$
\mat S = \begin{pmatrix}
 \xi & 0 \\
 0 & 1 - \sigma_{23}^2
\end{pmatrix},
$$%
we get%
\begin{align}
\Prob(X_1 = 1\mid Z_3 = z) % \hspace{-40pt}& \\
&=
  \int \phi(u;m_2, s_{22})
  \Phi\left(u; 0, \xi \right)\der u \nonumber\\
&= \Phi(0; - m_2, \xi + s_{22})
  =\Phi\left(
  \frac{\mu_2 + \sigma_{23}z}
  {\sqrt{1 + \xi - \sigma_{23}^2}}\right) \nonumber
\end{align}%
where the result is attained
using the identity in supplementary material  \ref{sec:CDFIdentity}.
This is the Gaussian copula model we have been working with if we let
$\xi\rightarrow 0^+$ which effectively means that $Z_1 = 0$.\footnote{%
To see this, compare
Equation \eqref{eqn:Z3DModX1} with $Z_1 = 0$ with Equation
\eqref{eqn:BinarySpec} and use $\mu_2 = \Phi^{-1}(p_1)$.}
In practice, we derive the formulas we need and work with $\xi = 0$ and we only
work with one latent variable for binary variables as \cite{zhao20}.

Now, consider the extension to the scenario with one multinomial variable
$X_1 \in \{0, \cdots, G - 1\}$,
and $L$ continuous variables. Assume that
$\vec Z \sim N^{(L + G)}(\vec\mu,\mat\Psi$) where %
\begin{align*}
\mat\Psi &= \begin{pmatrix}
    \xi & \vec 0^\top_{G - 1} & \vec 0^\top_L \\
    \vec 0_{G - 1} & \mat\Psi_{2:G2:G} & \mat\Psi_{2:G(-1:G)} \\
    \vec 0_L & \mat\Psi_{(-1:G)2:G} & \mat\Psi_{(-1:G)(-1:G)}
  \end{pmatrix} \\
\mat\Psi_{2:G2:G} &= \begin{pmatrix}
	1 & \sigma_{23} & \cdots & \sigma_{2K} \\
	\sigma_{32} & \sigma_{33} & \ddots & \sigma_{3G} \\
	\vdots & \ddots & \ddots &  \vdots	 \\
	\sigma_{G2} & \sigma_{G3} & \dots & \sigma_{GG}
  \end{pmatrix} \\
X_1 &= j -1 \Leftrightarrow
  Z_j \geq \max(Z_1,\dots, Z_G) \\
\vec X_{(-1)} &= \vec f^{-1}(\vec Z_{(-1:G)}),
\end{align*}%
$j \in \{1,\dots, G\}$,
$\vec\mu = (0, \vec\mu_{2:G}^\top, \vec 0_L^\top)^\top$,
$\mat\Psi_{(-1:G)2:G} \in\mathbb R^{L\times (G - 1)}$
is a dense matrix,
and $\mat\Psi_{(-1:G)(-1:G)} \in\mathbb R^{L\times L}$ is
a matrix in which diagonal entries are restricted to be one.\footnote{%
We arbitrarily select the first category to be the reference
and the corresponding $\vec Z$ entry is uncorrelated with all other
$\vec Z$ entries.}
This parameterization is used by \cite{bunch91},
in the absence of the continuous variables.
We have $G$ conditional probabilities given by  %
\begin{multline*}
\Prob(X_1 = j - 1 \mid  \vec Z_{(-1:G)} = \vec z) =
   \int\phi(u;m_j, s_{jj})
    \Phi^{(G - 1)}\bigg( \\
    \vec 1_{G - 1} u; \big(\vec m_{(-j)} + \mat S_{(-j)j}
    \frac{u - m_j}{s_{jj}}\big),
    \big(\mat S_{(-j)(-j)} - s_{jj}^{-1}\mat S_{(-j)j}\mat S_{j(-j)}\big)
    \bigg)\der u
\end{multline*}%
which equals %
\begin{multline*}
\Phi^{(G-1)}\bigg(\vec 0_{G - 1};
 \vec m_{(-j)} - \vec 1_{G - 1} m_j,\mat S_{(-j)(-j)}
  + s_{jj}\vec 1_{G - 1}\vec 1_{G - 1}^\top - \vec 1_{G - 1}\mat S_{j(-j)}
   - \mat S_{(-j)j}\vec 1_{G - 1}^\top
 \bigg)
\end{multline*}
where $\vec Z_{1:G} \mid \vec Z_{(-1:G)} = \vec z \sim N(\vec m, \mat S)$, %
\begin{align*}
\vec m &= \vec\mu_{1:G} +
  \mat\Sigma_{1:G(-1:G)}\mat\Sigma_{(-1:G)(-1:G)}^{-1}\vec z \\
\mat S &= \mat\Sigma_{1:G, 1:G} -
  \mat\Sigma_{1:G(-1:G)}\mat\Sigma_{(-1:G)(-1:G)}^{-1}
  \mat\Sigma_{(-1:G)1:G},
\end{align*}%
$\vec 1_k$ is a vector with $k$ ones, and
$(-j)$ or $(-\mathcal J)$ in a subscript
implies all elements but the $j^{\text{th}}$ or those indices in $\mathcal J$
for vectors or all but those indices in the rows or columns
for matrices.

\subsection{New Log Marginal Likelihood}\label{sec:NewLogML}
We can now turn to the general case with all types including
multiple multinomial variables and
show the log marginal likelihood term for each of the $n$ observations.
Suppose that the entries $X_{ij}$ with
$j \in \mathcal M \subseteq \{1,\dots,K\}$ are
multinomial variables
with
$\mathcal M \cap \mathcal C = \emptyset$,
$\mathcal M \cap \mathcal O = \emptyset$, and
$\mathcal B = \{1, \dots, K\}\setminus \mathcal C \cup \mathcal O \cup \mathcal M$.
Each variable $X_{ij}\in \{0,1,\dots m_j - 1\}$ has
$m_j$ categories when  $j \in \mathcal M$.
We permute the the observed variables,
the $\vec X_i$'s, such that the multinomial variables
are first without loss of generality and let
$c_{\mathcal M} = \lvert \mathcal M\rvert$ and
$c_{\setminus \mathcal M} = K - c_{\mathcal M}$.
The model is then
$
\vec Z_i \sim N^{(W)}\left(\vec\mu,
  \mat\Psi\right)
$
where $W = c_{\setminus \mathcal M} + \sum_{j = 1}^{c_{\mathcal M}} m_j$,
$\mat\Psi$ is given in supplementary material  \ref{sec:PsiMat},
and
$\lvert\mathcal B\rvert + \sum_{j = 1}^{c_{\mathcal M}}m_j - c_{\mathcal M}$
entries of $\vec\mu$ are non-zero.
One can permute $\mat\Psi$ with a permutation matrix $\mat P$ such
that:%
\begin{equation}\label{eqn:PsiParamatization}
\mat P \mat\Psi \mat P^\top = \begin{pmatrix}
    \xi\mat I_{c_{\mathcal M}} & \mat 0_{c_{\mathcal M} \times (W - c_{\mathcal M})} \\
    \mat 0_{(W - c_{\mathcal M}) \times c_{\mathcal M}} & \mat\Sigma
  \end{pmatrix}
\end{equation}%
where $\mat I_k$ is the $k$ dimensional identity matrix,
$\mat 0_{k\times l}$ is a $k\times l$ matrix with zeros, and
$\mat\Sigma$ is a symmetric positive definite matrix where all elements are free except for $K$ of the diagonal entries which are restricted to be one.
Thus, we parameterize $\mat\Psi$ in terms of a log Cholesky
decomposition of $\mat\Sigma$. As before, the restriction in the diagonal
of $\mat\Sigma$ is either applied after each stochastic gradient
iteration by scaling the rows and columns,
similar to what \cite{zhao20} do,
or by using a constrained nonlinear optimization method.

We now turn to the new expression of the log marginal likelihood.
Let %
$$
\mat D = \begin{pmatrix}
  \vec 1_{m_1 - 1} & \vec 0_{m_1 - 1} & \cdots & \vec 0_{m_1 - 1} \\
  \vec 0_{m_2 - 1} & \vec 1_{m_2 - 1} & \ddots & \vdots  \\
  \vdots & \ddots  & \ddots & \vec 0_{m_{c_{\mathcal M} - 1} - 1} \\
  \vec 0_{m_{c_{\mathcal M}} - 1} & \dots &
    \vec 0_{m_{c_{\mathcal M}} - 1} & \vec 1_{m_{c_{\mathcal M}} - 1} \\
  \vec 0_{\lvert\mathcal O \cup \mathcal B\rvert} &
    \cdot & \vec 0_{\lvert\mathcal O \cup \mathcal B\rvert} &
    \vec 0_{\lvert\mathcal O \cup \mathcal B\rvert}
\end{pmatrix}.
$$%
Furthermore, we let $\mathcal I_i$
be the set with the indices of the latent variables belonging to each of the
observed categories for the multinomial variables for observation
$i$. That is%
$$
\mathcal I_i = \left\{
  k = 1,\dots, c_{\mathcal M}:\, X_{ik} + 1 + \sum_{l = 1}^{k - 1} m_l
  \right\}
$$%
where $\sum_{l = 1}^0 m_l = 0$ by definition.

Let $\tilde{\mathcal C}$ be the indices of the latent
variables corresponding to the observed continuous variables and
similarly define $\tilde{\mathcal O}$, $\tilde{\mathcal B}$, and
 $\tilde{\mathcal M}$. Using %
\begin{align}
\bar{\vec\mu} &=
  \vec\mu_{\tilde{\mathcal B}\cup\tilde{\mathcal O}\cup
    \tilde{\mathcal M}}
  + \mat\Psi_{
  \tilde{\mathcal B}\cup\tilde{\mathcal O}\cup\tilde{\mathcal M},
  \tilde{\mathcal C}}
  \mat\Psi_{\tilde{\mathcal C}\tilde{\mathcal C}}^{-1}
  \widehat{\vec z}_{i\tilde{\mathcal C}}
  \label{eqn:condMean}\\
\bar{\mat S} &= \mat\Psi_{\tilde{\mathcal B}\cup\tilde{\mathcal O}
   \cup\tilde{\mathcal M},
 \tilde{\mathcal B}\cup\tilde{\mathcal O}\cup
    \tilde{\mathcal M}}  -
  \mat\Psi_{\tilde{\mathcal B}\cup\tilde{\mathcal O}
    \cup\tilde{\mathcal M},\tilde{\mathcal C}}
  \mat\Psi_{\tilde{\mathcal C}, \tilde{\mathcal C}}^{-1}
  \mat\Psi_{\tilde{\mathcal C},
    \tilde{\mathcal B}\cup\tilde{\mathcal O}\cup\tilde{\mathcal M}},
    \label{eqn:condVar}
\end{align} %
we can apply the identity in supplementary material  \ref{sec:CDFIdentity}
to get the following log marginal likelihood expression%
\begin{align}
l_i(\mat\Psi, \vec\mu) &=
  \log\phi^{(\lvert\mathcal C\rvert)}(
  \widehat{\vec z}_{i\tilde{\mathcal C}}; \vec 0,
  \mat\Psi_{\tilde{\mathcal C}\tilde{\mathcal C}}) +
  \log \int \phi^{(c_{\mathcal M})}(
  \vec u; \bar{\vec\mu}_{\mathcal I_i}, \bar{\mat S}_{
    \mathcal I_i\mathcal I_i}
  )\Phi^{(W)}
  \Big(\vec a_i,\vec b_i;
  \label{eqn:finalLogMl}\\
&\hspace{-10pt}
  \big(\bar{\vec\mu}_{(- \mathcal I_i)} - \mat D\vec u
  + \mat S_{(-\mathcal I_i)\mathcal I_i}
  \bar{\mat S}_{\mathcal I_i\mathcal I_i}^{-1}(\vec u -
  \bar{\vec\mu}_{\mathcal I_i})\big),
  \big(
  \bar{\mat S}_{(-\mathcal I_i)(-\mathcal I_i)} -
  \bar{\mat S}_{(-\mathcal I_i)\mathcal I_i}
  \bar{\mat S}_{\mathcal I_i\mathcal I_i}^{-1}
  \bar{\mat S}_{\mathcal I_i(-\mathcal I_i)}\big)\Big)\der\vec u
  \nonumber\\
&=
  \log\phi^{(\lvert\mathcal C\rvert)}(
  \widehat{\vec z}_{i\mathcal C}; \vec 0,
  \mat\Psi_{\tilde{\mathcal C}\tilde{\mathcal C}}) +
  \log\Phi^{(W)}
  \Big(  \nonumber\\
&\hspace{-10pt}
  \vec a_i,\vec b_i; \big(\bar{\vec\mu}_{(- \mathcal I_i)} -
    \mat D\bar{\vec\mu}_{\mathcal I_i}\big),
    \big(
    \bar{\mat S}_{(-\mathcal I_i)(-\mathcal I_i)}
  + \mat D\bar{\mat S}_{\mathcal I_i\mathcal I_i}\mat D^\top
  - \mat D\bar{\mat S}_{\mathcal I_i(-\mathcal I_i)}
  - \bar{\mat S}_{(-\mathcal I_i)\mathcal I_i}\mat D^\top
  \big)\Big).
  \nonumber
\end{align}

A missing $X_{ij}$ implies that the corresponding $\vec Z_i$
entry or entries are unrestricted. Thus, they
do not contribute
anything to the log marginal likelihood in Equation \eqref{eqn:orgLogML} and
\eqref{eqn:finalLogMl} after integrating them out
and can be omitted. Notice that the non-zero mean terms,
some of the terms of
$\vec\mu_{\tilde{\mathcal B}\cup\tilde{\mathcal O}\cup\tilde{\mathcal M}}$,
enter linearly in Equation \eqref{eqn:condMean}. This makes it easy to jointly
optimize the mean and covariance matrix, rather then fixing the mean as suggested by
\cite{zhao20},
as we have a gradient approximation of the mean of the \ac{CDF}.
Thus, we can easily estimate the mean vector for the binary and
multinomial variables together with $\mat\Sigma$.

\subsection{Application}\label{sec:multApplication}
We compare our method with other top performing
single imputation methods in this section.
The methods we compare with are missForest and the PCA-like
imputeFAMD. We are particularly interested in (i) the improvements from
using our method for a correctly specified model and
(ii) the performance of all three methods on observational data. We start with
(i) which will give an indication of the improvements we may expect when the
data generating process is approximately a Gaussian copula.

% latex table generated in R 4.1.0 by xtable 1.8-4 package
% Thu Jul  1 11:57:56 2021
\begin{table*}[t]
\centering
\caption{Average \acp{RMSE} and classification errors on simulated data sets and three observational data sets. The average computation times in seconds are given in the last column.} 
\label{tab:multRes}
\begingroup\tiny
\begin{tabular}{llS[table-format = 1.4(4)]S[table-format = 1.3(3)]S[table-format = 1.3(3)]S[table-format = 1.3(3)]|S[table-format = 2.2(2)]}
  \toprule
Data set & Method & {Binary} & {Ordinal} & {Continuous} & {Multinomial} & {Time} \\ 
  \midrule
Simulation & RQMC (our) & \bftab 0.3051(64) & \bftab 0.647(6) & \bftab 0.883(13) & \bftab 0.535(5) & 47.05(99) \\ 
   & missForest & 0.3451(65) & 0.692(6) & 0.949(11) & 0.603(6) & \bftab 8.72(25) \\ 
   & imputeFAMD & 0.3573(74) & 0.701(6) & 0.929(11) & 0.613(7) & 81.19(38) \\ 
   [.35em] Cholesterol & RQMC (our) & 0.3009(12) & \bftab 0.679(2) &  & \bftab 0.538(2) & 10.64(83) \\ 
   & missForest & 0.4022(46) & 0.696(3) &  & 0.700(8) & \bftab 2.88(12) \\ 
   & imputeFAMD & \bftab 0.2994(11) & 0.691(2) &  & 0.564(3) & 5.61(27) \\ 
   [.35em] Rent & RQMC (our) & \bftab 0.1096(7) & \bftab 0.370(4) & \bftab 0.731(3) & \bftab 0.890(2) & 173.55(54) \\ 
   & missForest & 0.1408(9) & 0.389(4) & 0.757(4) & 0.903(2) & \bftab 19.07(70) \\ 
   & imputeFAMD & 0.1121(8) & 0.537(5) & 0.777(3) & 0.959(2) & 102.97(224) \\ 
   [.35em] Colon & RQMC (our) & \bftab 0.2217(14) & \bftab 0.223(2) & 1.029(8) & 0.659(4) & 28.73(91) \\ 
   & missForest & 0.2666(25) & 0.394(5) & 1.060(7) & \bftab 0.614(4) & \bftab 11.50(44) \\ 
   & imputeFAMD & 0.2275(14) & 0.223(2) & \bftab 1.017(7) & 0.671(4) & 51.42(98) \\ 
   \bottomrule
\end{tabular}
\endgroup
\end{table*}

For (i) and (ii)
we generate $100$ data sets and
within each there is a 30 percent chance that
a variable is missing (missing independently at random).
The imputeFAMD has a tuning parameter which is the number of components. We
estimate this using five-fold cross validation.
In our simulation study (i) we simulate data from
a Gaussian copula in a similar manner to that described in Section
\ref{sec:compWOMult}. That is, we simulate
$\mat\Sigma$ in $\mat\Psi$ in Equation \eqref{eqn:PsiParamatization} from a
Wishart distribution like in the previous simulation. We have $8$
variables per observation with $2$ of each type. The ordinal
and multinomial variables have five equally likely categories and there is a
total 2000 observations.

The result of the simulation study is shown in the ``Simulation'' rows in Table
\ref{tab:multRes}.\footnote{%
We do not include the \ac{SMAE} as there is no obvious way
to compute this for the multinomial variables, the integer values assigned
to each category of the ordinal variables is arbitrary,
and it favors the Gaussian copula models for
the ordinal variables as the other methods treat ordinal
variables as multinomial variables (especially if we use the
conditional median similar to what \cite{zhao20} do as shown in Table
\ref{tab:simZhaoError}).}
Our method performs much better for all four types of variables.
The average computation times for the simulated and observational data show
that our method is the slowest of the three.\footnote{%
We use four threads with missForest and imputeFAMD does not support
computation in parallel.}
However, it is not orders of magnitude slower.

We include three observational data sets: the rent data set, used also in Section
\ref{sec:compWOMult}, but this time with the multinomial area code,
the cholesterol data from a US survey from the survey package \citep{Lumley20},
and data from one of the first successful trials of adjuvant chemotherapy for
colon cancer from the survival package \citep{Therneau20}. Our method often
performs best or is close to the best performing method. It is expected that
our method is not always the best performing method as we do make a parametric
assumption with the particular copula we use. It is, however, encouraging that
the data generating processes seem well approximated with the Gaussian copula for
all three data sets.

\section{Conclusion and Discussion}\label{sec:conclusion}
We have shown that direct maximum likelihood estimation of the
model used by \cite{zhao20,zhao20Mat} using \ac{RQMC} is feasible
and provides improvements compared with the approximation that
they use. Our \ac{RQMC} method yielded
lower errors for the imputed values, and estimated model parameters
closer to the true values in our simulation study.
Importantly, the error of our method decreases faster as a function of the
number of samples.
We have also extended the model
to support multinomial variables which increases the applicability of the
method.

\subsection{Future Work}

Our
\ac{RQMC} method can be extended
to provide e.g.\ an arbitrarily precise approximation of the quantiles
of the latent variables conditional on the observed data and estimated
model parameters.
A major advantage of having precise quantile approximations is that the
user can get a quantile
estimate at a given level
which is close to the nominal level if the data generating
process is well approximated by the Gaussian copula.
These may provide substantial improvements over the lower bounds and
approximate confidence intervals based on Gaussian approximations,
that \cite{zhao20Mat} suggest using, since the conditional distribution of the
latent variables may be very non-Gaussian.

We could replace the non-parametric density estimator with a flexible parametric
transformation like those used in transformation models \cite{Hothorn18}.
This would lead to a fully parametric model for which it would be possible to do joint
optimization over all parameters, including the borders for the ordinal
variables. This could substantially increase the efficiency and subsequently the
performance. Our \ac{RQMC} could be extended to use the recently developed
minimax tilting method by \cite{Botev17} which could decrease the
computation time at a fixed variance of the estimator.
Finally, our method can easily be used for multiple imputation
(to, importantly, account for imputation uncertainty),
since it is a probabilistic framework.

\bibliographystyle{apa}
\bibliography{ref}

% \appendix
\clearpage
\setcounter{section}{0}
\renewcommand{\thesection}{S\arabic{section}}

\section{Multivariate Normal CDF Identity}\label{sec:CDFIdentity}
In this section, we show an identity which we will use repeatably. Let %
$$\label{eqn:exGenSkew}
\begin{pmatrix}
  \vec V_1 \\ \vec V_2
\end{pmatrix} \sim
  N^{(k_1 + k_2)}\left(
  \begin{pmatrix}
  \vec \xi_1 \\ \vec\xi_2
  \end{pmatrix},
  \begin{pmatrix}
  \mat\Xi_{11} & \mat\Xi_{12} \\
  \mat\Xi_{21} & \mat\Xi_{22}
  \end{pmatrix}
  \right)
$$%
where $\vec V_1\in\mathbb{R}^{k_1}$ and $\vec V_2\in\mathbb{R}^{k_2}$,
$\vec\xi_1$ and $\vec\xi_2$ are mean vectors for $\vec V_1$ and $\vec V_2$,
respectively,
and $\mat\Xi$ is a covariance matrix where the sub matrices have $k_1$ or $k_2$
rows and columns.
Then the joint density of $\vec V_1 = \vec v_1$ and $\vec a < \vec V_2 < \vec b$
(a box constraint on $\vec V_2$) is%
\begin{align*}
\phi^{(k_1)}(\vec v_1; \vec \xi_1, \mat\Xi_{11})
  \Prob\left(\vec a < \vec V_2 < \vec b \,\middle\vert\,
  \vec V_1 = \vec v_1
  \right)\nonumber
  \hspace{-180pt}& \\
&= \phi^{(k_1)}(\vec v_1; \vec \xi_1, \mat\Xi_{11})
  \Phi^{(k_2)}\bigg(
  \vec a, \vec b;
  \big(\vec \xi_2 + \mat\Xi_{21}\mat\Xi_{11}^{-1}(\vec v_1 - \vec\xi_1)\big),
  \big(\mat\Xi_{22} - \mat\Xi_{21}\mat\Xi_{11}^{-1}\mat\Xi_{12}
  \big)\bigg)
\end{align*}%
and the marginal for $\Prob(\vec a < \vec V_2 < \vec b)$ is %
\begin{align*}
\Prob(\vec a < \vec V_2 < \vec b) \hspace{-60pt}\\
&=
  \Phi^{(k_2)}(\vec a, \vec b; \vec\xi_2, \mat\Xi_{22})  \\
%  \label{eqn:margSkewCondLHS}
&=
  \int \phi^{(k_1)}(\vec v_1; \vec \xi_1, \mat\Xi_{11})
  \Prob\left(\vec a < \vec V_2 < \vec b \,\middle\vert\,
  \vec V_1 = \vec v_1
  \right)\der\vec v_1.
%  \label{eqn:margSkewCondRHS}
\end{align*}%
Next, we can define %
\begin{align*}
\mat Z &= \mat\Xi_{21}\mat\Xi_{11}^{-1} \\
\vec c &= \vec\xi_2 - \mat\Xi_{21}\mat\Xi_{11}^{-1}\vec\xi_1
  = \vec\xi_2 - \mat Z\vec\xi_1 \\
\mat\Omega &= \mat\Xi_{22} - \mat\Xi_{21}\mat\Xi_{11}^{-1}\mat\Xi_{12}
  = \mat\Xi_{22} - \mat Z\mat\Xi_{11}\mat Z^\top
\end{align*} %
which we can use to show that%
\begin{equation}\label{eqn:CDFIdentity}
\int \phi^{(k_1)}(\vec x;\vec\mu,\mat\Sigma)
  \Phi^{(k_2)}(\vec a,\vec b;\vec c + \mat Z\vec x, \mat\Omega)\der\vec x
  =
  \Phi^{(k_2)}(\vec a,\vec b; \vec c + \mat Z\vec\mu,
  \mat\Omega + \mat Z\mat\Sigma\mat Z^\top).
\end{equation}

\section{Parameterization for the General Model with More than One Multinomial Variable}\label{sec:PsiMat}

Following the notation in Section \ref{sec:NewLogML}, we can generalize
the covariance matrix for $\vec Z$
in Section \ref{sec:addMult} to include all variable types including
multiple multinomial variables by letting $\mat\Psi$ be:

\begin{align}
\mat\Psi &= \begin{pmatrix}
    \mat \Psi^{(11)} & \cdots & \mat \Psi^{(1l)} \\
    \vdots & \ddots & \vdots \\
    \mat\Psi^{(l1)} & \dots  &
      \mat\Psi^{(ll)}
  \end{pmatrix}
  \nonumber\\
\mat \Psi^{(kk)} &= \begin{pmatrix}
    \xi & 0 & 0 & \cdots & 0 \\
	0 & 1 & \psi^{(kk)}_{23} & \cdots & \psi^{(kk)}_{2m_k} \\
	0 & \psi^{(kk)}_{32} & \psi^{(kk)}_{33} & \ddots & \psi^{(kk)}_{3m_k} \\
	\vdots & \vdots & \ddots & \ddots &  \vdots	 \\
	0 &\psi^{(kk)}_{m_k2} & \psi^{(kk)}_{m_k3} & \dots & \psi^{(kk)}_{m_km_k}
  \end{pmatrix}
  \nonumber\\
\mat \Psi^{(kk')} &= \begin{pmatrix}
    0 & 0 & \cdots & 0 \\
	0 & \psi^{(kk')}_{22}  & \cdots & \psi^{(kk')}_{2m_{k'}} \\
	\vdots & \vdots & \ddots &  \vdots	 \\
	0 &\psi^{(kk')}_{m_k2}  & \dots & \psi^{(kk')}_{m_km_{k'}}
  \end{pmatrix}
  \nonumber\\
\mat \Psi^{(kl)} &=
  \begin{pmatrix}
    0 & \cdots & 0 \\
    \psi^{(kl)}_{21} & \cdots &
      \psi^{(kl)}_{2c_{\setminus\mathcal M}} \\
   \vdots
     & \ddots & \vdots \\
   \psi^{(kl)}_{m_k1} & \cdots &
      \psi^{(kl)}_{m_kc_{\setminus\mathcal M}}
  \end{pmatrix}
  \nonumber\\
\mat \Psi^{(ll)} &= \nonumber\\
&\hspace{-20pt}
  \begin{pmatrix}
    1 & \psi^{(ll)}_{12} & \cdots
      & \psi^{(ll)}_{1c_{\setminus\mathcal M}} \\
    \psi^{(ll)}_{21} & 1 & \ddots
      & \vdots \\
    \vdots & \ddots & \ddots &
      \psi^{(ll)}_{c_{\setminus\mathcal M} - 1, c_{\setminus\mathcal M}}
     \nonumber\\
     \psi^{(ll)}_{c_{\setminus\mathcal M}1} &
       \dots &
        \psi^{(ll)}_{c_{\setminus\mathcal M}, c_{\setminus\mathcal M} - 1} & 1
  \end{pmatrix}
  \nonumber
\end{align}%
for $k,k' \in \{1,\dots, c_{\mathcal M}\}\wedge k'\neq k$ and
$l = c_{\mathcal M} + 1$. The $k^{\text{th}}$
$\mat \Psi^{(kk)}$ in the diagonal for $k = 1,\dots,c_{\mathcal M}$
is for the latent variables for the $k^{\text{th}}$ multinomial variable.
The last $\mat \Psi^{(ll)}$ block
is for the latent variables for the
binary, continuous, and ordinal variables.

\section{Derivative Approximations}\label{sec:Derivs}
We show the gradient of the log marginal likelihood with respect
to the covariance matrix, $\mat\Psi$, in this section.
Without loss of generality,
suppose that covariance matrix $\mat\Psi$ is permutated
such that the first indices are for the latent variables for the
continuous variables, the next  $\lvert\mathcal M\rvert$
indices are the latent variables corresponding to the observed categories
of the multinomial variables, and the final indices are the remaining
latent variables for the multinomial, ordinal, and binary variables.
That is, %
\begin{align*}
\mathcal R_i &= \tilde{\mathcal M} \cup \tilde{\mathcal O}
  \cup \tilde{\mathcal B} \setminus \mathcal I_i \\
\mat\Psi &= \begin{pmatrix}
  \mat\Psi_{\mathcal C\mathcal C} &
    \mat\Psi_{\mathcal C\mathcal I_i} &
    \mat\Psi_{\mathcal C\mathcal R_i} \\
  \mat\Psi_{\mathcal I_i\mathcal C} &
    \mat\Psi_{\mathcal I_i\mathcal I_i} &
    \mat\Psi_{\mathcal I_i\mathcal R_i} \\
   \mat\Psi_{\mathcal R_i\mathcal C} &
     \mat\Psi_{\mathcal R_i\mathcal I_i} &
     \mat\Psi_{\mathcal R_i\mathcal R_i}
\end{pmatrix} \\
\bar{\vec\mu} &= (
  \bar{\vec\mu}_{1:c_{\mathcal M}},
  \bar{\vec\mu}_{(-1:c_{\mathcal M})})^\top \\
\bar{\mat S} &=
  \begin{pmatrix}
    \bar{\mat S}_{1:c_{\mathcal M}1:c_{\mathcal M}} &
     \bar{\mat S}_{1:c_{\mathcal M}(-1:c_{\mathcal M})} \\
    \bar{\mat S}_{(-1:c_{\mathcal M})1:c_{\mathcal M}} &
     \bar{\mat S}_{(-1:c_{\mathcal M})(-1:c_{\mathcal M})}
  \end{pmatrix}
\end{align*} %
where  $\bar{\vec\mu}$ and $\bar{\mat S}$ are given in
Equation \eqref{eqn:condMean} and \eqref{eqn:condVar}.
It follows that the log marginal likelihood in Equation
\eqref{eqn:finalLogMl} is%
\begin{align*}
l_i &=
  \log\phi^{(\lvert\mathcal C\rvert)}(
  \widehat{\vec z}_{i\mathcal C}; \vec 0,
  \mat\Psi_{\mathcal C\mathcal C})
  \log\Phi^{(c_{\setminus \mathcal M} +
  \sum_{j = 1}^{c_{\mathcal M}} (m_j - 1))}
  \left(\vec a_i,\vec b_i; \vec m,\mat M\right) \\
\vec m &=
  \bar{\vec\mu}_{(-1:c_{\mathcal M})} -
    \mat D\bar{\vec\mu}_{1:c_{\mathcal M}} \\
\mat M &=
  \bar{\mat S}_{(-1:c_{\mathcal M})(-1:c_{\mathcal M})} +
  \mat D\bar{\mat S}_{1:c_{\mathcal M}1:c_{\mathcal M}}\mat D^\top
  - \mat D\bar{\mat S}_{1:c_{\mathcal M}(-1:c_{\mathcal M})}
  - \bar{\mat S}_{(-1:c_{\mathcal M})1:c_{\mathcal M}}\mat D^\top.
\end{align*}

For a matrix $\mat A \in \mathbb R^{g\times g}$, we let%
$$
\mat A' = \begin{pmatrix}
  \partial l_i / \partial a_{11}  & \cdots &
    \partial l_i / \partial a_{1g} \\
  \vdots & \ddots & \vdots \\
  \partial l_i / \partial a_{g1} & \dots &
    \partial l_i / \partial a_{gg}
\end{pmatrix}.
$$%
It then follows that%
\begin{align*}
\bar{\mat S}' &=
  \begin{pmatrix}
    \mat D^\top \mat M'\mat D & -\mat D^\top \mat M' \\
    -\mat M'\mat D & \mat M'
  \end{pmatrix} \\
\mat\Psi_{\mathcal I_i\cup\mathcal R_i, \mathcal I_i\cup\mathcal R_i}'
  &= \bar{\mat S}' \\
\bar{\vec\mu}' &=
  \begin{pmatrix}
    \mat D^\top\vec m' \\
    \vec m'
  \end{pmatrix} \\
\mat\Psi_{\mathcal C\mathcal C}' &=
  \mat\Psi_{\mathcal C\mathcal C}^{-1}
  \mat\Psi_{\mathcal C,\mathcal I_i\cup\mathcal R_i}\bar{\mat S}'
  \mat\Psi_{\mathcal I_i\cup\mathcal R_i, \mathcal C}
  \mat\Psi_{\mathcal C\mathcal C}^{-1}
  -\frac 12\mat\Psi_{\mathcal C\mathcal C}^{-1}
  \mat\Psi_{\mathcal C,\mathcal I_i\cup\mathcal R_i}
  \bar{\vec\mu}'
  \widehat{\vec z}_{i\mathcal C}^\top
  \mat\Psi_{\mathcal C\mathcal C}^{-1} \\
&\hspace{10pt}-\frac 12\mat\Psi_{\mathcal C\mathcal C}^{-1}
  \widehat{\vec z}_{i\mathcal C}\bar{\vec\mu}^{\prime\top}
  \mat\Psi_{\mathcal I_i\cup\mathcal R_i,\mathcal C}
  \mat\Psi_{\mathcal C\mathcal C}^{-1}
  -\frac 12\mat\Psi_{\mathcal C\mathcal C}^{-1} + \frac 12
  \mat\Psi_{\mathcal C\mathcal C}^{-1}
  \widehat{\vec z}_{i\mathcal C}\widehat{\vec z}_{i\mathcal C}^\top
  \mat\Psi_{\mathcal C\mathcal C}^{-1} \\
\mat\Psi_{\mathcal I_i\cup\mathcal R_i,\mathcal C}' &=
  -\bar{\mat S}'
  \mat\Psi_{\mathcal I_i\cup\mathcal R_i, \mathcal C}
  \mat\Psi_{\mathcal C\mathcal C}^{-1}
  + \frac 12\bar{\vec\mu}'
  \widehat{\vec z}_{i\mathcal C}^\top
  \mat\Psi_{\mathcal C\mathcal C}^{-1}.
\end{align*}

Thus, we can get an approximation of all the derivatives we need
if we have an approximation of $\vec m'$ and
$\mat M'$. Details of our approximation of the latter
quantities are provided in Section \ref{sec:newMethod} and in the next section.

\section{Quasi-Monte Carlo Procedure}\label{sec:QMCMethod}

\begin{algorithm}[ht!]
\DontPrintSemicolon

  \KwInput{Lower and upper bounds, $\vec a\in \mathbb R^k$ and $\vec b \in \mathbb R^k$,
           mean vector $\vec\omega\in \mathbb R^k$,
           covariance matrix $\mat \Omega\in\mathbb R^{k\times k}$,
           number of samples $S$,
           function $\vec g:\, \mathbb R^k\rightarrow \mathbb R^H$,
           and procedure to generate a $k$
           dimensional quasi-random sequence or pseudorandom numbers in
           $(0,1)^k$}
  \KwOutput{Approximation of $\int_{a_1}^{b_1}\cdots\int_{a_k}^{b_k}
 \vec g(\vec u)\phi^{(k)}(\vec u;\vec\omega,\mat\Psi)\der \vec u$}
  Compute the Cholesky decomposition $\mat O^\top\mat O =\mat \Sigma$
  and set $\vec r = \vec 0_H$ \\
  \For{$j=1$ \KwTo $k$}{
     Set $a_j\leftarrow o_{jj}^{-1}(a_j - \omega_j)$,
     $b_j\leftarrow o_{jj}^{-1}(b_j - \omega_j)$, and
     $\vec o_{1:j,j}\leftarrow o_{jj}^{-1}\vec o_{1:j,j}$
  }
  \For{$s=1$ \KwTo $S$}{
    Draw the next $\vec u \in (0,1)^k$ and set $w =1 $ \\
    \For{$j=1$ \KwTo $k$}{
      Compute
\begin{align*}
\hat a_j &= \begin{cases}
   a_1 & j = 1 \\
   a_j - \vec o_{1:(j - 1),j}^\top\vec x_{1:(j - 1)} & j > 1
  \end{cases} \\
\hat b_j &= \begin{cases}
   b_1 & j = 1 \\
   b_j - \vec o_{1:(j - 1),j}^\top\vec x_{1:(j - 1)} & j > 1
\end{cases}
\end{align*}
      Set $w\leftarrow w \cdot (\Phi(\hat b_j) -\Phi(\hat a_j))$ and
      $x_j = \Phi^{-1}(\Phi(\hat a_j) + u_j (\Phi(\hat b_j) -\Phi(\hat a_j)))$
    }
    Update the mean estimator, $\vec r \leftarrow \vec r + s^{-1}(w \vec g(\mat O^\top\vec x + \vec\omega) - \vec r)$
  }

  \Return $\vec r$

\caption{(Quasi) \ac{MC} procedure to approximate integrals that are
needed to
estimate the model and impute the missing values.}\label{alg:QMCMethod}
\end{algorithm}

Algorithm \ref{alg:QMCMethod} shows pseudocode for the method we
use to approximate the intractable integrals in the
log likelihood, the gradient of the log likelihood, and the quantities used
to perform the imputation. The algorithm is $\bigO{k^3}$ because of the
Choleksy decomposition but the primary bottleneck for practical problems is the
loop. For small to moderate $k$ (say $k < 50$),
the computation time spent evaluating $\Phi$
and $\Phi^{-1}$ is substantial. For moderate to large $k$, the dot product
in $\hat a_j$ and $\hat b_j$ takes a relatively larger part of the computation
time.

The dot product in $\hat a_j$ and $\hat b_j$ does not take full advantage of
\ac{SIMD} instructions on modern CPUs.
Thus, we found substantial reductions in
computation time by simultaneously processing multiple draws.
An adaptive \ac{RQMC} method can be used
if we run algorithm \ref{alg:QMCMethod} using multiple randomized
quasi-random sequences
in parallel as \cite{Genz02}. This allows us to get an estimate of the error
which can be used to stop early if the error is less than a user specified
threshold.
We use the Fortran code written by \cite{Genz02} to simultaneously compute the
Cholesky decomposition and find the permutation of the variables.
The permutation is based on
a heuristic to reduce the variance when we approximate the likelihood with
$\vec g(\vec x) = 1$.

As for the gradient, let%
$$
\tilde L(\vec\omega, \mat\Omega) = \int_{a_1}^{b_1}\cdots\int_{a_k}^{b_k}
 \phi^{(k)}(\vec u;\vec\omega,\mat\Omega)\der \vec u.
$$%
Then the gradient of the likelihood is given%
\begin{align*}
\nabla_{\vec\omega}\tilde L(\vec\omega, \mat\Omega) &=
  \mat\Omega^{-1}\int_{a_1}^{b_1}\cdots\int_{a_k}^{b_k}
  (\vec u - \vec\omega)
  \phi^{(k)}(\vec u;\vec\omega,\mat\Omega)\der \vec u \\
&= \mat O^{-1}\int \vec u h(\vec u)\prod_{j = 1}^k w_j(\vec u)\der u \\
\nabla_{\mat\Omega}\tilde L(\vec\omega, \mat\Omega) &= \\
&\hspace{-10pt}
  \frac 12 \left(\mat\Omega^{-1}\left(
  \int_{a_1}^{b_1}\cdots\int_{a_k}^{b_k}
  (\vec u - \vec\omega)(\vec u - \vec\omega)^\top
  \phi^{(k)}(\vec u;\vec\omega,\mat\Omega)\der\vec u
  \right)\mat\Omega^{-1} - \mat\Omega^{-1}
  \tilde L(\vec\omega, \mat\Omega)\right) \\
&= \frac 12 \left(
  \mat O^{-1}\left(
  \int \vec u\vec u^\top h(\vec u)\prod_{j = 1}^k w_j(\vec u)\der u
  \right)\mat O^{-\top} - \mat\Omega^{-1}
  \tilde L(\vec\omega, \mat\Omega)\right)
\end{align*}%
with $\nabla_{\mat\Omega} = (\partial / \partial\Omega_{ij})_{i,j=1,\dots, k}$
and $h$ is given by the importance distribution in Equation
\eqref{eqn:importDist}.
The needed choice of $\vec g$ can be seen from the two integrals above but it
can also be seen that one can simplify the expressions by working with
$\vec g(\mat O^\top\vec x + \vec\omega)$ in Algorithm \ref{alg:QMCMethod}.

% latex table generated in R 4.1.0 by xtable 1.8-4 package
% Thu Jul  1 11:57:56 2021
\begin{table*}[t]
\centering
\caption{Summary statistics for each observational data set. The first column is the number observations. The other columns indicates the number of variables of each type except for the ``levels'' columns which show the maximum number of categories for the ordinal and multinomial variables, respectively.} 
\label{tab:dataStats}
\begingroup\tiny
\begin{tabular}{lrrrrrrr}
  \toprule
 & \# Observations & Binary & Continuous & Ordinal & Levels & Multinomial & Levels \\ 
  \midrule
Medcare & 4406 &   2 &   2 &   4 &  19 &   0 &   0 \\ 
  Rent & 2053 &   7 &   3 &   1 &   6 &   1 &  25 \\ 
  ESL & 488 &   0 &   0 &   5 &  10 &   0 &   0 \\ 
  LEV & 1000 &   0 &   0 &   5 &   5 &   0 &   0 \\ 
  GBSG & 686 &   3 &   6 &   1 &   3 &   0 &   0 \\ 
  TIPS & 244 &   3 &   2 &   2 &   6 &   0 &   0 \\ 
  Cholesterol & 7846 &   2 &   0 &   1 &   4 &   1 &   4 \\ 
  Colon & 1776 &   5 &   2 &   2 &   4 &   1 &   3 \\ 
   \bottomrule
\end{tabular}
\endgroup
\end{table*}

\section{Data Sets}\label{sec:dataInfo}
We will briefly describe the data sets we have used in this section. We have
removed incomplete observations from each data set. We removed the
potentially right censored survival time along with the censoring status from
the colon data set.
Such variables can be properly handled by extending our imputation method to
support censored variables, which at present are not supported.
This data set is not used by \cite{zhao20} and, therefore, there are not any
results to compare with.

In contrast, we keep the
potentially right censored survival
time and the censoring status for the \ac{GBSG} data set as do \cite{zhao20}.
Thus, our results can be compared with their results.

Table \ref{tab:dataStats} summarises the number of variables of each type,
the maximum number of categories for the ordinal and multinomial variables, and
the numbers of observations in the complete data sets.

\end{document}